\documentclass[preprint,12pt]{elsarticle}

\usepackage{amssymb}

\usepackage{times}  
\usepackage{helvet}  
\usepackage{courier}  
\usepackage[hyphens]{url}  
\usepackage{graphicx}
\usepackage{natbib}  
\usepackage{caption}

\usepackage{algorithm}
\usepackage{algorithmic}
\usepackage{booktabs}       
\usepackage{amsfonts}       
\usepackage{nicefrac}       
\usepackage{microtype}      
\usepackage{xcolor}         
\usepackage{graphicx}
\usepackage{amsthm,amsmath,amssymb}
\usepackage{mathrsfs}
\usepackage{bbm}
\usepackage{dsfont}
\usepackage{amsmath, amsfonts, amssymb}
\usepackage{makecell}
\usepackage{mathrsfs}
\usepackage{stfloats}
\usepackage{amsmath}
\usepackage{newfloat}
\usepackage{listings}

\journal{*}
\begin{document}

\begin{frontmatter}



\title{Complementary Labels Learning with Augmented Classes}

\author[a,d]{Zhongnian Li}
\author[a]{Jian Zhang}
\author[c]{Mengting Xu}
\author[a]{Xinzheng Xu$^{2,}$}
\author[b,d]{Daoqiang Zhang}
\address[a]{organization={School of Computer Science and Technology, China University of Ming and Technogy},
	city={Xuzhou},
	postcode={221000}, 
	state={Jiangsu},
	country={China}}
\address[b]{organization={College of Computer Science and Technology, Nanjing University of Aeronautics and Astronautics},
	city={Nanjing},
	postcode={210000}, 
	state={Jiangsu},
	country={China}}
\address[c]{organization={college of computer science and technology, Zhejiang University},
	city={Hangzhou},
	postcode={310000}, 
	state={Zhejiang},
	country={China}}
\address[d]{organization={MIIT Key Laboratory of Pattern Analysis and Machine Intelligence},
	city={Nanjing},
	postcode={210000}, 
	state={Jiangsu},
	country={China}}

\fntext[fn2]{This work is supported by the National Natural Science Foundation of China (No.61976217), the Fundamental Research Funds for the Central Universities(No.2019XKQYMS87), the Opening Fundation of Key Laboratory of Opto-technology and Intelligent Control (Lanzhou Jiaotong University), Ministry of Education (KFKT2020-3) and the Science, Technology Planning Project of Xuzhou (No.KC21193) and the Fundamental Research Funds for the Central Universities( No.NJ2022028).}
\fntext[fn1]{Corresponding author}

\begin{abstract}
Complementary Labels Learning (CLL) arises in many real-world tasks such as private questions classification and online learning,  which aims to alleviate the annotation cost compared with standard supervised learning. 
Unfortunately, most previous CLL algorithms were in a stable environment rather than an open and dynamic scenarios, where data collected from  unseen augmented classes in the training process might emerge in the testing phase. 
In this paper, we propose a novel problem setting called  Complementary Labels Learning with Augmented Classes (CLLAC), which brings the challenge that classifiers trained by complementary labels  should not only be able to classify  the  instances from observed classes accurately, but also recognize the instance from the Augmented Classes in the testing phase. Specifically, by using unlabeled data, we propose an unbiased  estimator of classification risk for CLLAC, which is guaranteed to be provably consistent. Moreover, we provide generalization error bound for proposed method which shows that the optimal parametric convergence rate is achieved for estimation error.   Finally, the experimental results on  several benchmark datasets verify the effectiveness of the proposed method. 

\end{abstract}

\begin{keyword}
Complementary Labels, Augmented Classes, Unbiased  Estimator 

\end{keyword}

\end{frontmatter}


\section{Introduction}
Ordinary machine learning task often requires large-scale data with accurate label information \cite{DBLP:conf/aaai/XueH19}, which is costly and time-consuming. To mitigate this problem, the weakly supervised learning \cite{DBLP:journals/pami/GongYYS22} paradigm has been investigated, which allows  machine learning  techniques to work  with less expensive data. In this paper, we focus on a challenging setting of weakly supervised learning called Complementary Labels Learning (CLL) \cite{DBLP:conf/nips/IshidaNHS17,DBLP:conf/eccv/YuLGT18, DBLP:conf/icml/FengK000S20, DBLP:conf/aaai/WangWPZ21, DBLP:conf/ijcai/WangFZ21}, which annotates one of classes that an instance does not belong to. Since complementary labels are less informative than supervised ordinary labels \cite{DBLP:journals/ijcv/TianXYYL22},  it is obvious that obtaining data with complementary label is less expensive and time-consuming than supervised multi-class label. Another potential application of CLL is data privacy \cite{DBLP:conf/nips/IshidaNHS17}, which evades the private answers by specifying the incorrect label.

Previous methods of CLL usually modified the supervised ordinary classification 	risk estimator and surrogate loss function to learn from complementary labels. 
\cite{DBLP:conf/nips/IshidaNHS17} propose the unbiased risk estimator of the classifier, which shows solid theoretical foundations.  \cite{DBLP:conf/eccv/YuLGT18} address  CLL with the biased complementary labels, and propose a  forward loss correction method to learn the true labels.  \cite{DBLP:conf/icml/FengK000S20} introduce multiple complementary labels setting, which specifies multiple classes that the instance doses not belong to.  \cite{DBLP:conf/aaai/XuGCLZB20} introduced the  complementary conditional Generative Adversarial Nets to model instance distribution for learning the complementary labels classifier. In this paper, we deal with the complementary labels with unbiased assumption.

\begin{figure*}
	\centering
	\includegraphics[width=0.78\textwidth]{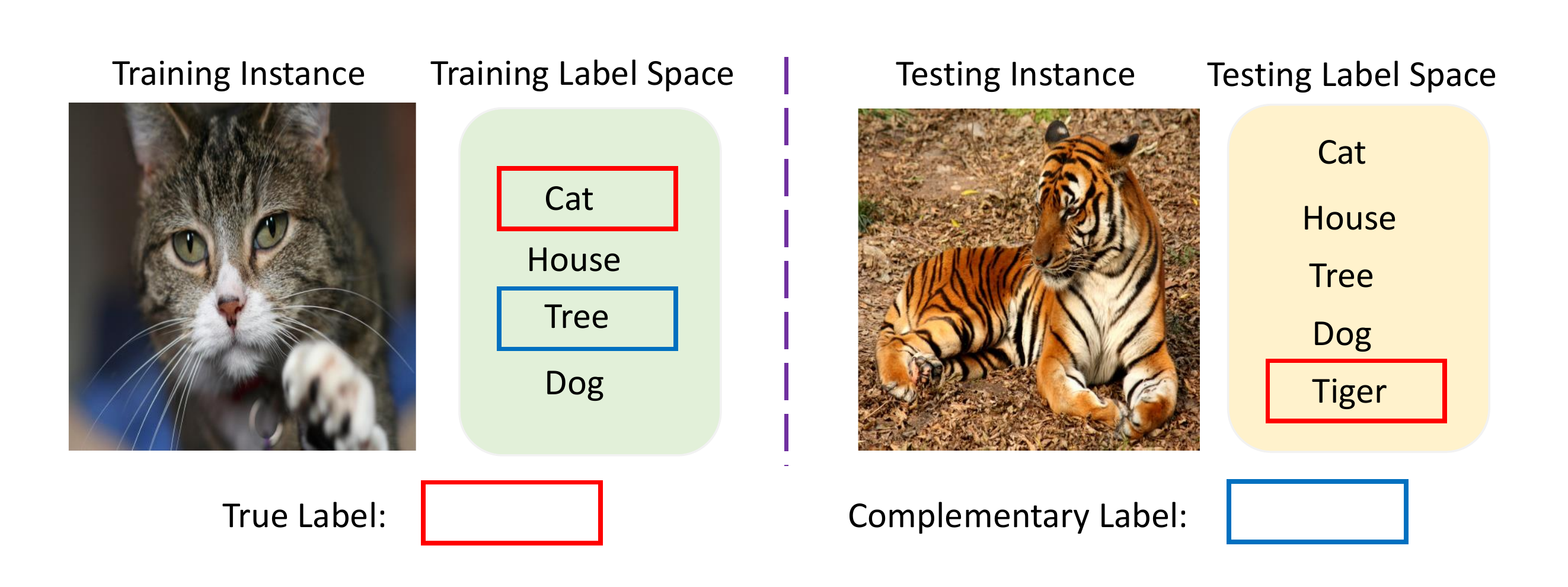}
	\caption{One example of complementary labels data with  Augmented Classes. The augmented class (Tiger) only appear in testing dataset, which is not observed in complementary labels data.}
	\label{motivation}
	\vspace{-1em}
\end{figure*}

Unfortunately, recent CLL \cite{DBLP:conf/icml/Chou0LS20, DBLP:conf/icml/GaoZ21} methods are in a stationary environments rather than an environment that is open and changes gradually, which faces many challenges raised in real-world classification \cite{DBLP:journals/tkde/WeiYMWSZ21, DBLP:conf/aaai/DaYZ14}. For example, in classification on the Internet, the user may only annotates few classes by complementary label: the training data contains four classes, e.g., cat, house, tree and dog. However, the classifier has to predict augmented classes in the future, such as tiger in Fig.\ref{motivation}, which might make the classifier unusable. When faced with these real-world applications, the augmented classes need to be considered for a mature CLL method.  

In this paper,  we focus on Complementary Labels Learning with unseen Augmented Classes (CLLAC), which breaks the traditional stationary setting of supervised learning \cite{DBLP:journals/tkde/WeiYMWSZ21, DBLP:conf/aaai/0006ZZ21, DBLP:conf/aaai/DaYZ14, DBLP:conf/icml/Zhang0JZ20}.  In augmented classes classification setting \cite{DBLP:conf/aaai/DaYZ14, DBLP:conf/nips/Zhang0MZ20},  augmented classes are unknown during the training stage, but emerge in the test stage. The goal of augmented classes classification is not only to recognize the instance from the observed classes labeled in the training stage accurately, but also to classify the instance in the augmented classes.  The main challenge of the CLLAC lies under that it is difficult to model relationships between known complementary labels and augmented classes data. 

In order to deal with above problem, this paper gives the first attempt to propose an Unbiased risk estimator for  Complementary Labels Learning with Augmented Classes (UCLLAC). More concretely, by exploiting unlabeled training data, which was collected easily during the training stage in many real-life applications, unbiased risk of classifier over testing distribution can be rewritten in the training stage. Our unbiased risk estimator builds a prototype baseline for CLLAC that might inspire more studies for this new problem in the future. Then,  we provide a theoretical analysis of estimation error bound, which justifies the capability of our method in exploiting unlabeled data. In our experiments, we show that our method significantly outperforms the state-of-the-art methods on benchmark datasets including MNIST, fashion-MNIST, Kuzushi-MNIST and CIFAR-10. The experimental results verify that our method not only accurately recognizes the instance from the observed classes labeled in the training stage, but also classifies the instance in the augmented classes.

\section{Preliminaries}
In this section, we first review some formulations of complementary labels learning, then introduce some notations for classification with augmented classes.
\subsection{Complementary Labels Learning}
In multi-class classification, suppose the feature space is $\mathcal{X} \in \mathbb{R}^d$ and $\mathcal{Y} = \{1,2,...,k\}$ is label space, where $d$ is the feature space dimension and $k$ is the number of classes. The labeled sample $(x,y)\in \mathcal{X} \times \mathcal{Y} $  is sampled from an unknown probability density $p(x,y)$.  Ordinary classification arms to learn an classifier $f(x):\mathbb{R}^d \to \mathbb{R}^k$ by minimizing the expected risk as follows:
\begin{equation}
	R(f) = {\mathbb{E}_{(x,y) \sim p(x,y)}}[l(f(x),y)]	
\end{equation}where $l(f(x),y)$ denotes the multi-class loss function and $\mathbb{E}_{(x,y) \sim p(x,y)}$ is the expectation over density $p(x,y)$.  Since the density $p(x,y)$ is unknown, we usually approximate expected risk $R(f)$ by using its empirical estimation.  

Next, in the complementary labels learning setting \cite{DBLP:conf/nips/IshidaNHS17},  suppose labeled  sample  $(x,\overline y )\in \mathcal{X} \times \mathcal{Y} $ is sampled from complementary labels data density $\overline p(x,\overline y)$, where $\overline y \in \mathcal{Y}$  denotes the complementary labels of instance $x$. Then, the complementary labels data density can be expressed as follows:  
\begin{equation}
	\overline p (x,\overline y ) = \frac{1}{{k - 1}}\sum\nolimits_{y \ne \overline y } {p(x,y)} 
	\label{cll_eq}
\end{equation}which implies that the generative assumption is unbiased. Since $\overline p(x) = p(x)$ , according to Eq.\ref{cll_eq}, we have $\overline p (\overline y \left| x \right.) = \frac{1}{{k - 1}}\sum\nolimits_{y \ne \overline y } {p(y\left| x \right.)} $, which naturally lead to an unbiased risk estimator to rewrite the ordinary classification risk estimator. By using one-versus-all(OVA) strategy, the risk of complementary labels learning can be formulated  as follows:  
\begin{equation}
	\overline R (f) = (k - 1){\mathbb{E}_{(x,\overline y ) \sim \overline p (x,\overline y )}}[\overline L (f(x),\overline y )] - {M_1} + {M_2}
\end{equation}where ${M_1} = \sum\limits_{\overline y  = 1}^k {\overline L (f(x),y)} $,  ${M_2} = \overline L (f(x),y) + L(f(x),y)$,  $ L (f(x),y) = l({f_y}(x)) + \frac{1}{{k - 1}}\sum\limits_{y' \ne y} {l( - {f_{y'}}(x))} $ and $\overline L(f(x), \overline y) = \frac{1}{{k - 1}}\sum\limits_{y' \ne \overline y } {l({f_{y'}}(x))}  + l( - {f_{\overline y }}(x))$. For this risk, if binary loss $l(z)$ satisfies $l(z)+l(-z) = 1$, then $M_1 = k$ and $M_2 =2$, which implies that the method has restriction on the loss functions.

In order to overcome above problem, \cite{DBLP:conf/icml/IshidaNMS19} proposed an general unbiased risk estimator, which allows to use the popular loss, such as softmax cross-entropy. The general risk of complementary labels learning can be formulated as follows:
\begin{equation}
	\overline R (f) ={\mathbb{E}_{(x,\overline y ) \sim \overline p (x,\overline y )}}[ - (k - 1)L(f(x),\overline y ) + \sum\limits_{y = 1}^k {L(f(x),y)} ]	
\end{equation}where $L(f(x),y)$ denotes the OVA loss function.  It is worth noting that this method can use any loss for complementary labels learning.

Although feasible CLL approaches have been proposed for many real-world applications by exploiting the CLL generative assumption, they were in a stable environment rather than scenarios that  opens and changes gradually. 

\subsection{Learning with Augmented Classes}
Traditional multi-class classification faces many challenges raised in real-life applications, where the open and  dynamic scenarios break the stationary setting   implied in traditional classification \cite{DBLP:journals/tkde/WeiYMWSZ21, DBLP:conf/aaai/0006ZZ21, DBLP:conf/aaai/DaYZ14, DBLP:conf/icml/GuoZJLZ20, DBLP:conf/icml/Zhang0JZ20}. Learning with Augmented Classes  (LAC) \cite{DBLP:conf/aaai/DaYZ14, DBLP:conf/nips/Zhang0MZ20, DBLP:conf/icdm/DingLZ18} is one branch method to deal with the dynamic environments, which aims to handle new classes emerging in the learning process by exploiting unlabeled data (including augmented classes). Another similar problem is Open Set Recognition (OSR) \cite{DBLP:journals/pami/GengHC21}, which is set up with no unlabeled data available in the training stage. 

While some studies have focused on the LAC, there is no work concerning complementary labels learning with augmented classes yet, which might make unreliable predictions.

\section{Methodology}
In this section, we study how to learn complementary labels with augmented classes. We first introduce the problem formalization of CLLAC.  Then we describe our unbiased risk estimator for CLLAC. Thirdly, we further derive the theoretical analysis of proposed approach.
\subsection{Problem Formalization}
In the training stage, the learning system collects a complementary labels dataset $D_{CL} = \{(x_i,\overline {y_i})\}_{i=1}^{n_{kcl}}$ sampled from  known complementary labels data density  $\overline p_{kcl}(x,\overline y)$ defined over $ \mathcal{X} \times \mathcal{Y}$. In the testing stage, learning system requires to predict instance sampled from the testing density $p_{te}(x,y)$, where augmented classes not observed in the complementary labels dataset might emerge. Since the number of augmented classes is unknown, the learning system will classify all of augmented classes as a single augmented class $ac$. Then, we can obtain the definition of testing density over $ \mathcal{X} \times \mathcal{Y}_{ac}$, where $\mathcal{Y}_{ac} = \{1,...,K,ac\}$ is the augmented class space. In LAC setting,  besides the complementary labels dataset, the learning system additionally collects a set of unlabeled data $D_u = \{x_i\}_{i=1}^{n_u}$, which was sampled from the testing density $p_{te}(x,y)$. The goal of CLLAC is to learn a model $f(x_i)\to y_i \in \mathcal{Y}_{ac}$, which might achieve good generalization ability over the testing distribution.

In the CLLAC problem, the  unlabeled  samples are a mixture of samples in $K+1$ classes, therefore the testing density  can be denoted by a linear combination of complementary labels data density  and 
the density of augmented classes. In this paper, we assume that the distribution of known complementary labels data remains unchanged when augmented classes emerge. Then, the testing density $p_{te}$ can be expressed as:
\begin{equation}
	{p_{te}} = \theta {\overline p _{kcl}} + (1 - \theta ){p_{ac}}
\end{equation}where $p_{ac}$ denotes the density of augmented classes and $\theta \in [0,1]$ denotes a certain mixture proportion. Based on this expression, we can learn the classifier by minimizing the classification risk over the testing distribution.
\subsection{Unbiased Risk Estimator}
In this section, we propose an unbiased risk estimator to learn from complementary labels with augmented classes data.  In the rest of this paper, we assume complementary labels data is sampled from complementary density, i.e., Eq.\ref{cll_eq}, which makes the complementary labels problem  degenerates to standard multi-class classification. Then, the CLLAC problem can be addressed by rewriting the risk, when the joint testing distribution is available. 

In the ordinary classification, given the joint distribution and adopted the one-versus-rest (OVR) strategy, the risk is formulated as follows: 
\begin{equation}\label{or_risk}
	\begin{split}
		{R_\phi }({f_1},...,{f_{K+1}}) &= \mathbb{E}_{(x,y) \sim {p_{te}}}[\phi ({f_y}(x)) \\ & + \sum\nolimits_{k = 1,k \ne y}^{K + 1} \phi ( - {f_k}(x))] 
	\end{split}
\end{equation}where $f_k$ denotes the classifier for the $k-$th class, $f_{K+1}$ denotes the classifier for augmented classes, $\phi$ denotes the binary surrogates loss for ordinary classification. The OVR strategy predicts label by $f(x) = argmax_{i \in{1,...,K+1}}f_i(x)$

\textbf{Rewriting the risk.} Unfortunately, the joint testing density is unavailable in the training stage due to the only existing complementary labels data and the absence of labeled data from augmented classes. Suppose the mixture proportion is available, the density of augmented classes data can be evaluated by separating the density of complementary labels data from the unlabeled data as 
\begin{equation}
	(1 - \theta ){p_{ac}}(x) = {p_{te}}(x) - \theta {\bar p_{kcl}}(x)
\end{equation}
Then, the risk of ordinary classification can be approximated with the complementary labels and unlabeled data as follows:

\textbf{Lemma 1.} Let $F$ denotes $(f_1,...,f_{K+1})$ and $f_i$ denotes the classifier for the $i-$th class, the classification risk ${R_{CLLAC} }({f_1},...,{f_{K+1}})$ can be equivalently represented as
\begin{equation}\label{lemma1}
	\begin{split}
		R_{\phi}(F) &= R_{CLLAC}(F)  =  {{\theta \mathbb{E}}_{(x,\bar y) \sim {{\bar p}_{kcl}}}} [- (K - 1)l(\bar y,F(x)) \\&+ \sum_{j = 1}^K {l(j,F(x))} ] + {{\mathbb{E}}_{{\rm{x}} \sim {p_{te}}(x)}}[l(K + 1,F(x))] \\&- \theta {\mathbb{E}_{{\rm{x}} \sim {p_{kcl}}(x)}}[l(K + 1,F(x))]
	\end{split}
\end{equation} where $p_{te}$ and $p_{kcl}$ denote the instance densities of testing and complementary labels distributions respectively. $l(j,F(x))$ denotes the loss as follows:
\begin{equation}\label{loss}
	l(j,F(x)) = \phi ({f_j}) + \sum\limits_{k = 1,k \ne j}^{K + 1} {\phi ( - {f_k})} 
\end{equation}
Thus, by plugging Eq.\ref{loss} into Eq.\ref{lemma1}, we can rewrite the ordinary classification by using the complementary labels and unlabeled densities in the training stage as follows:

\textbf{Proposition 2.} Let $f_k$ denotes the classifier for the $k-$th class, the risk of CLLAC can be  equivalently represented as:
\begin{equation}
	\begin{split}
		{R_{CLLAC}(F)} = &\theta {\mathbb{E}_{(x,\bar y) \sim {{\bar p}_{kcl}}}}[(\phi ( - {f_{K + 1}}) - \phi ({f_{K + 1}}))\\& + (K - 1)[\phi ( - {f_{\bar y}}) - \phi ({f_{\bar y}})]] \\&+ {\mathbb{E}_{x \sim {p_{te}}(x)}}[\phi ({f_{K + 1}}) + \sum\limits_{j = 1}^K {\phi ( - {f_j})} ]
	\end{split}
\end{equation}

Proof can be found in Appendix. Note that the risk of CLLAC is the non-convexity caused by terms $- \phi ({f_{K + 1}})$ and $- \phi ({f_{\bar y}})$ which  are non-convex w.r.t the classifier. In practice,  we can eliminate the non-convex by using some surrogate losses  which satisfy $\phi ( - z) - \phi (z) = -z$ for all $z \in\mathbb{R}$ \cite{DBLP:conf/icml/PlessisNS15}. Then, the convex formulation of  $R_{CLLAC}$ can be represented as follow:
\begin{equation}\label{convex}
	\begin{split}
		{R_{CLLAC}} = &\theta {\mathbb{E}_{(x,\bar y) \sim {{\bar p}_{kcl}}}}[-f_{K+1} - (K - 1)f_{\bar y}] \\&+ {\mathbb{E}_{x \sim {p_{te}}(x)}}[\phi ({f_{K + 1}}) + \sum\limits_{j = 1}^K {\phi ( - {f_j})} ]
	\end{split}
\end{equation} Given the linear classifiers, minimization the convex risk can  obtain the globally optimal solution. Many loss functions satisfy the convex condition, such as square loss $\phi(z) = \frac{(1-z)^2}{{{4}}}$ and logistic loss $\phi(z) = log(1+exp(-z))$. Since  the risk of ordinary classification is rewritten by the risk of CLLAC, its empirical estimator is unbiased over testing density.

According to Proposition 2, given the complementary labels and unlabeled dataset $D_{CLLAC}=\{(x_i,\overline {y_i})\}_{i=1}^{n_{kcl}} \cup \{x_i\}_{i=1}^{n_u}$, we can  obtain empirical approximation  of the unbiased risk estimator as follows:
\begin{equation}
	\begin{split}
		{\hat R_{CLLAC}} &=  \frac{\theta}{{{n_{kcl}}}} \sum\limits_{i = 1}^{{n_{kcl}}} [(\phi ( - {f_{K + 1}}({x_i})) - \phi ({f_{K + 1}}({x_i}))) \\&+ (K - 1)[\phi ( - {f_{{{\bar y}_i}}}({x_i})) - \phi ({f_{{{\bar y}_i}}}({x_i}))]]  \\&+  \frac{1}{{{n_{u}}}}\sum\limits_{i = 1}^{{n_u}} {[\phi ({f_{K + 1}}({x_i})) + \sum\limits_{j = 1}^K {\phi ( - {f_j}({x_i}))} ]} 
	\end{split}
\end{equation}

\textbf{Practical implementation.} We investigate implementation of our approach when using the deep neural networks. We choose square loss to formulate the unbiased risk estimator and minimize the non-convex risk of CLLAC to train the classifiers. In this way, the final objective function is not convex, which may inspire more specially optimization methods for this new setting to avoid the overfitting. Besides, in our method, many Mixture Proportion Estimation (MPE) works \cite{DBLP:conf/icml/RamaswamyST16} can be employed for estimating the mixture proportion $\theta$ directly.  
\subsection{Theoretical Analysis}
In this section, we first show the consistent analysis for the proposed method. Then, we provide the generalization error bound for unbiased risk estimator.

\textbf{Infinite-sample consistency.} Now, we derive the CLLAC risk $R_{CLLAC}$ is consistent with ordinary supervised risk  with respect to 0-1 loss.   The following theorem states that we can obtain the classifiers that achieve the Bayes rule by minimizing the rewritten risk  $R_{CLLAC}$.

\textbf{Theorem 3.} Let $f_1,...f_K,f_{K+1}$ denote binary classifiers for known labels and augmented classes respectively and $f(x)=argmax_{i\in\{1,...,K,K+1\}}f_i(x)$. Suppose the surrogate loss $\phi(z)$ is convex, bound below, differentiable, and for $z>0$, $\phi(z)<\phi(-z)$, $R_{0-1} = \mathbb{E}_{(x,y)\sim p_{te}}[\mathds{1}(f(x)\neq y )]$ denotes the supervised ordinary risk with respect to 0-1 loss, then for any $\epsilon_1 > 0$, there exists $\epsilon_2 > 0$ such that 
\begin{equation}
	\begin{split}
		R_{CLLAC}(f_1&,...,f_K,f_{K+1}) \leqslant R_{CLLAC}^ *  + {\varepsilon _2} \\& \Longrightarrow {R_{0 - 1}}(f) \leqslant {R_{0-1}^ * } + {\varepsilon _1}
	\end{split}
\end{equation}where $\mathds{1}(\bullet)$ denotes the indicator function,  $R_{CLLAC}^ * = min_{f_1,...,f_K,f_{K+1}}R_{CLLAC}(f_1,...,f_K,f_{K+1})$ and $R_{0-1}^ * = min_{f}R_{0-1}(f)= R_{0-1}(f^*)$ denotes the Bayes error for ordinary supervised  distribution. 

Theorem 3 states that we can obtain the Bayes classifiers which  achieve optimal classification error over the testing distribution. Since the CLLAC risk $R_{CLLAC}$ is consistent with ordinary supervised risk $R_{0-1}$, we can obtain  the same well-behaved classifiers  as learning with ordinary supervised data.

\textbf{Finite-sample generalization error bound.} Now, we derive the generalization error bound for unbiased risk estimator implemented by deep models with OVR strategy. Let $F=(f_1,...,f_{K+1})$ denotes classification vector function in hypothesis set $\mathcal{H}$ of the deep neural network. Suppose the surrogate loss $sup_{z}\phi(z)\leqslant C_{\phi}$, for $C_{\phi}>0$ and $L_{\phi}$ denotes the Lipschitz constant of $\phi$, we can derive the following lemma.

\textbf{Lemma 4.} For any $\delta>0$, with the probability at least $1-\delta/2$, we have
\begin{equation}
	\begin{split}
		{\sup _{F \in \mathcal{H}}}| & {{R_{kcl}}(F)-  {\widehat{R}_{kcl}}  (F)} | \\& \leqslant  4{K}{L_\phi }{\mathfrak{R}_{{n_{kcl}}}}(\mathcal{H}) + 2{K}{C_\phi }\sqrt {\frac{{\ln (4/\delta )}}{{2{n_{kcl}}}}} 
	\end{split}	
\end{equation}
\begin{equation}
	\begin{split}
		{\sup _{F \in \mathcal{H}}}&| {{R_{u}}(F) -  {\widehat{R}_{u}}  (F)} | \\& \leqslant 2{(K+1)}{L_\phi }{\mathfrak{R}_{{n_{kcl}}}}(\mathcal{H}) + {(K+1)}{C_\phi }\sqrt {\frac{{\ln (4/\delta )}}{{2{n_{u}}}}} 
	\end{split}
\end{equation} where $K$ denotes the number of known classes, $R_{kcl}(F) ={\mathbb{E}_{(x,\bar y) \sim {{\bar p}_{kcl}}}}[(\phi ( - {f_{K + 1}}) - \phi ({f_{K + 1}})) + (K - 1)[\phi ( - {f_{\bar y}}) - \phi ({f_{\bar y}})]]$ and $R_u(F) = {\mathbb{E}_{x \sim {p_{te}}(x)}}[\phi ({f_{K + 1}}) + \sum\limits_{j = 1}^K {\phi ( - {f_j})}] $, ${\widehat{R}_{kcl}}  (F)$ and  ${\widehat{R}_u}  (F)$ denote the empirical risk estimator to  $R_{kcl}(F)$ and $R_u(F)$ respectively, $\mathfrak{R}_{{n_{kcl}}}(\mathcal{H})$ and $\mathfrak{R}_{{n_u}}(\mathcal{H})$ are the Rademacher complexities \cite{DBLP:books/daglib/0034861} of $\mathcal{H}$ for the sampling of size $n_{kcl}$ from $\overline p_{kcl}(x,\overline y)$ and the sampling of size $n_u$ from $p_{te}(x,y)$. In this lemma, the hypothesis set $\mathcal{H}$ can be specified to some model hypothesis set, such as kernel-based  and deep-based classifiers.

According to lemma 4, we can derive the estimation error bound as follows.

\textbf{Theorem 5.} For any $\delta>0$, with the probability at least $1-\delta/2$, we have
\begin{equation*}
	\begin{split}
		R_{CLLAC}&({\hat {F}_{CLLAC})} -  \mathop{\rm {min}} _{{F} \in \mathcal{H}}R_{CLLAC}(F) \leqslant  \\& 8{K}{L_\phi }{\mathfrak{R}_{{n_{kcl}}}}(\mathcal{H}) + 4{(K+1)}{L_\phi }{\mathfrak{R}_{{n_u}}}(\mathcal{H})\\&+ 4{K}{C_\phi }\sqrt {\frac{{\ln (4/\delta )}}{{2{n_{kcl}}}}}+ 2{(K+1)}{C_\phi }\sqrt {\frac{{\ln (4/\delta )}}{{2{n_u}}}}
	\end{split}
\end{equation*} where $\hat {F}_{CLLAC}$ is trained by minimizing the CLLAC risk $R_{CLLAC}$

Theorem 5 shows that the proposed unbiased estimator exists an error bound which achieve the optimal convergence rate \cite{DBLP:journals/tit/Mendelson08}. With a growing number of complementary labels data and unlabeled data, the estimation error decreases. When the used deep neural network hypothesis set $\mathcal{H}$ is fixed and  $\mathfrak{R}_{{n}}(\mathcal{H}) \leqslant C_{\mathcal{F}}/\sqrt{n}$ \cite{DBLP:journals/chinaf/GaoZ16}, where $C_{\mathcal{F}}$ denotes the constant for weight and feature norms, we have $\mathfrak{R}_{{n_{kcl}}}(\mathcal{H}) = \mathcal{O}(1/\sqrt{n_{kcl}})$ and  $\mathfrak{R}_{{n_u}}(\mathcal{H}) = \mathcal{O}(1/\sqrt{n_u})$, then 
\begin{equation*}
	\begin{split}
		{n_{kcl}},{n_u}& \to \infty  \Longrightarrow \\&R_{CLLAC}({\hat {F}_{CLLAC})} - \mathop{\rm {min}} _{{F} \in \mathcal{H}}R_{CLLAC}(F)  \to 0
	\end{split}
\end{equation*} It is worth noting that when the number of classes $K$ increases,  the learning task becomes more difficult,  which is in line with our intuition.

\begin{table*}
	\centering
	\caption{Classification accuracy of each algorithm on benchmark datasets, with varying classes and the number of examples. (CLs), A means that classes $(1, 2,3), 4 $ are taken as complementary labels and augmented classes respectively.  We report the mean and standard deviation of results over 3 trials. The best method is shown in bold (under 5$\%$ t-test). }
	{\resizebox{0.96\textwidth}{45mm}{
			\renewcommand\arraystretch{1.15}
			\tabcolsep=0.7cm
			\begin{tabular}{lc|ccc cc }
				\toprule
				Dataset&(CLs), A&EUAC&UREA&CL-PC&CL-GA&UCLLAC\\
				\cmidrule{1-7} 
				&(0,1,2), 3&30.32$\pm$1.56&41.11$\pm$6.46&58.94$\pm$6.27&75.03$\pm$0.10&\textbf{97.73$\pm$0.52}\\ 
				&(1,2,3), 5&31.23$\pm$5.92&33.17$\pm$4.25&55.68$\pm$3.73&77.01$\pm$0.21&\textbf{96.33$\pm$0.58}\\ 
				&(1,2,3), 9&31.58$\pm$2.19&35.12$\pm$6.15&58.30$\pm$8.79&75.03$\pm$0.06&\textbf{97.06$\pm$0.44}\\ 
				MNIST&(4,5,6), 2&31.67$\pm$1.27&31.40$\pm$0.80&44.70$\pm$4.83&72.06$\pm$0.12&\textbf{95.55$\pm$1.07}\\
				&(4,5,6), 3&28.76$\pm$0.47&30.19$\pm$1.06&40.34$\pm$2.33&72.73$\pm$0.06&\textbf{96.22$\pm$0.37}\\
				&(6,7,8,9), 0&32.48$\pm$0.14&32.18$\pm$1.44&34.02$\pm$2.13&70.05$\pm$0.20&\textbf{95.32$\pm$0.19}\\ 
				&(7,8,9), 1, 2 &43.78$\pm$3.45&50.29$\pm$1.44&41.04$\pm$6.08&57.20$\pm$0.14&\textbf{95.44$\pm$0.32}\\ 
				\cmidrule{1-7} 				
				&(1,2,3), 0&32.17$\pm$3.25&36.03$\pm$5.60&36.39$\pm$6.60&72.15$\pm$0.15&\textbf{92.54$\pm$0.18}\\ 
				&(1,2,3), 9&38.59$\pm$7.61&42.71$\pm$8.89&45.32$\pm$5.48&72.27$\pm$0.19&\textbf{96.29$\pm$0.23}\\
				&(1,2,3), 8&31.67$\pm$4.35&44.47$\pm$5.15&41.90$\pm$9.09&72.23$\pm$0.06&\textbf{95.72$\pm$0.24}\\ 
				Fashion&(4,5,6), 9&33.19$\pm$3.23&31.95$\pm$2.48&44.87$\pm$4.16&68.18$\pm$0.21&\textbf{89.67$\pm$0.73}\\
				&(4,5,6), 3&31.52$\pm$1.80&32.10$\pm$2.07&49.76$\pm$2.32&69.10$\pm$0.15&\textbf{86.60$\pm$0.97}\\
				&(1,2,3,4), 0&25.4083$\pm$1.53&26.30$\pm$6.33&28.02$\pm$5.34&69.72$\pm$0.91&\textbf{87.34$\pm$1.29}\\
				&(7,8,9), 1, 2&44.28$\pm$2.58&57.49$\pm$0.34&35.54$\pm$7.89&58.41$\pm$0.07&\textbf{96.97$\pm$0.48}\\
				\cmidrule{1-7} 				
				&(1,2,3), 0&29.63$\pm$0.55&31.95$\pm$0.55&37.52$\pm$2.86&67.95$\pm$0.05&\textbf{86.44$\pm$0.86}\\ 
				&(1,2,3), 9&27.22$\pm$1.65&29.10$\pm$0.85&40.16$\pm$7.64&67.60$\pm$0.97&\textbf{85.11$\pm$0.70}\\
				&(1,2,3), 8&26.59$\pm$0.89&28.14$\pm$1.71&35.93$\pm$1.94&67.74$\pm$0.37&\textbf{83.77$\pm$1.31}\\ 
				Kuzushi&(4,5,6), 9&30.36$\pm$4.17&32.60$\pm$2.47&39.35$\pm$3.46&68.13$\pm$0.33&\textbf{85.16$\pm$1.02}\\
				&(4,5,6), 3&32.43$\pm$0.44&35.43$\pm$2.97&37.26$\pm$4.85&67.63$\pm$0.51&\textbf{86.03$\pm$0.10}\\
				&(4,5,6), 0&31.17$\pm$0.82&33.20$\pm$0.41&37.65$\pm$6.01&68.56$\pm$0.53&\textbf{84.35$\pm$1.35}\\ 
				&(7,8,9), 1&26.93$\pm$0.56&26.67$\pm$0.84&34.70$\pm$0.10&67.40$\pm$0.07&\textbf{80.14$\pm$1.57}\\
				\bottomrule
	\end{tabular}}}
	\label{tab_acc_mnist}
	\vspace{-1em}	
\end{table*}
\section{Experiments}
In this section, we evaluate our method from three aspects: (Q1) performance of recognizing  known classes and classifying augmented classes; (Q2) robustness for inaccurate training class priors; (Q3) capability of handling class probabilities shift. In the experiments, classifiers are trained with complementary labels data and unlabeled data, and we evaluate these classifiers on an additional testing data set which is never  collected in the training stage.
\begin{figure}
	\vspace{-0em}
	\centering
	\includegraphics[width=0.52\textwidth]{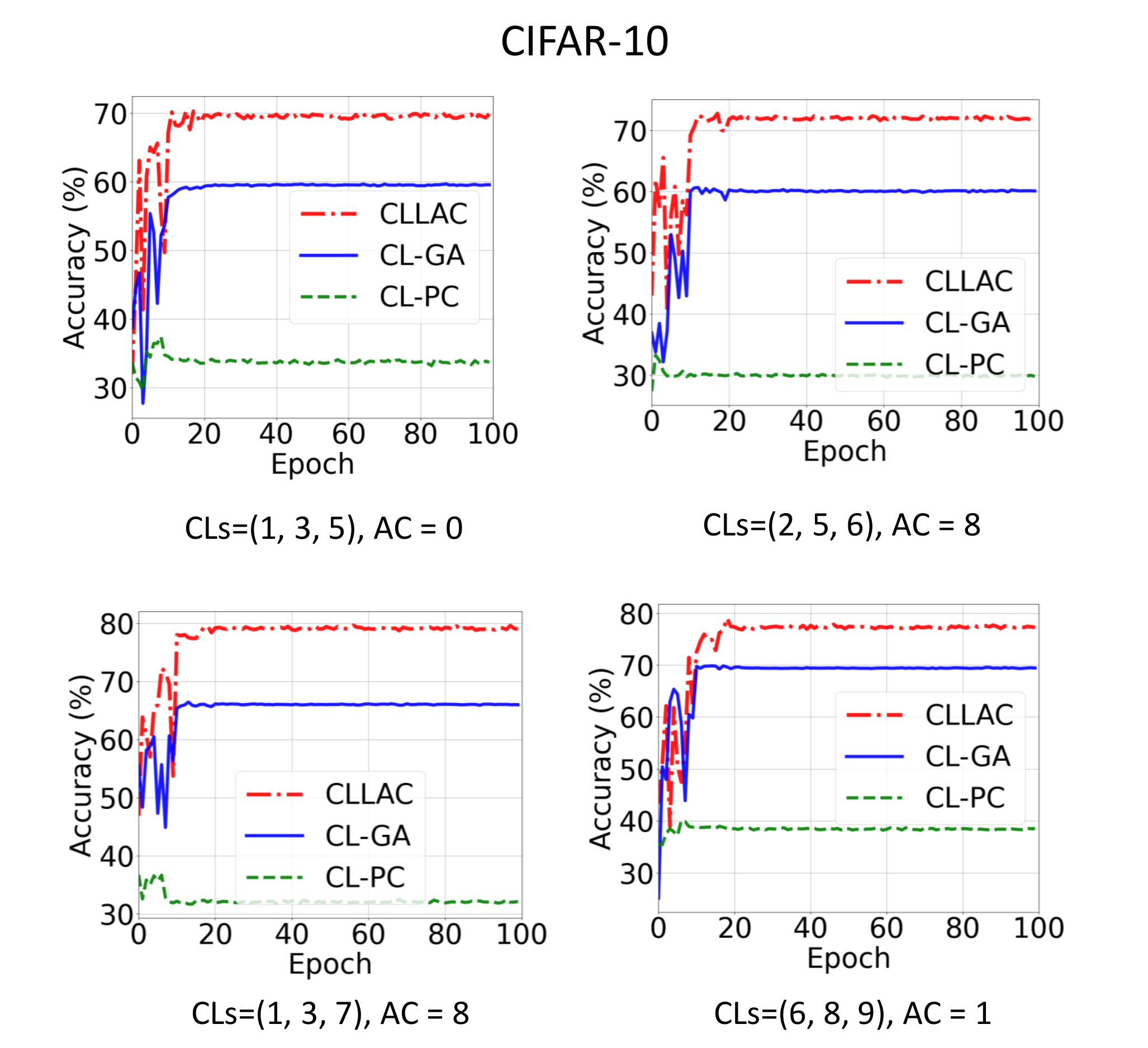}
	\caption{Illustrations classification accuracy of each complementary labels learning method on benchmark dataset CIFAR-10, with varying classes when the training epoch increases. CLs $=(1, 3, 5)$ means that classes 1, 3, 5 are taken as complementary labels. AC $=0$ means that class 0 is taken as augmented class. }
	\label{results_cifar}
	\vspace{2em}
\end{figure}
\subsection{Experiments Setup}
\textbf{Datasets:} We conduct experiments on four benchmark datasets: MNIST, Fasion-MNIST, Kuzushi-MNIST and CIFAR-10.  MNIST dataset is a widely-used benchmark dataset which consists of 10 classes. with 60000 training samples and 10000 testing sampling. The sizes of  Fashion-MNIST, Kuzushi-MNIST and CIFAR-10 are similar to MNIST.  We construct the complementary labels and augmented classes datasets as follows: we first select three and one classes from original datasets  as complementary labels and augmented classes datasets respectively. Then, we randomly select samples from complementary labels dataset and samples from augmented classes dataset as unlabeled dataset. Besides,  we generate complementary label by randomly specifying one of class which the instance dose not belong to. 

\textbf{Comparison Methods:} We compare the proposed approach with state-of-the-art methods from several diverse domain including: \textbf{EUAC} \cite{DBLP:conf/nips/Zhang0MZ20} is a state-of-the-art method for multi-class classification with augmented classes, which exploits unlabeled training data to enable classifier to identify the augmented classes. \textbf{UREA} \cite{DBLP:conf/icdm/ShuLY020} is a powerful Multi-Class Positive and Unlabeled (MPU) method, which is similar to multi-class learning with augmented classes. \textbf{CL-PC} \cite{DBLP:conf/nips/IshidaNHS17} is a powerful scheme for complementary labels learning, which uses ramp loss to train classifiers.  \textbf{CL-GA} \cite{DBLP:conf/icml/IshidaNMS19} is anther state-of-the-art complementary labels learning that allows to train a classifier with any loss.
\begin{figure*}
	\vspace{-0em}
	\centering
	\includegraphics[width=0.94\textwidth]{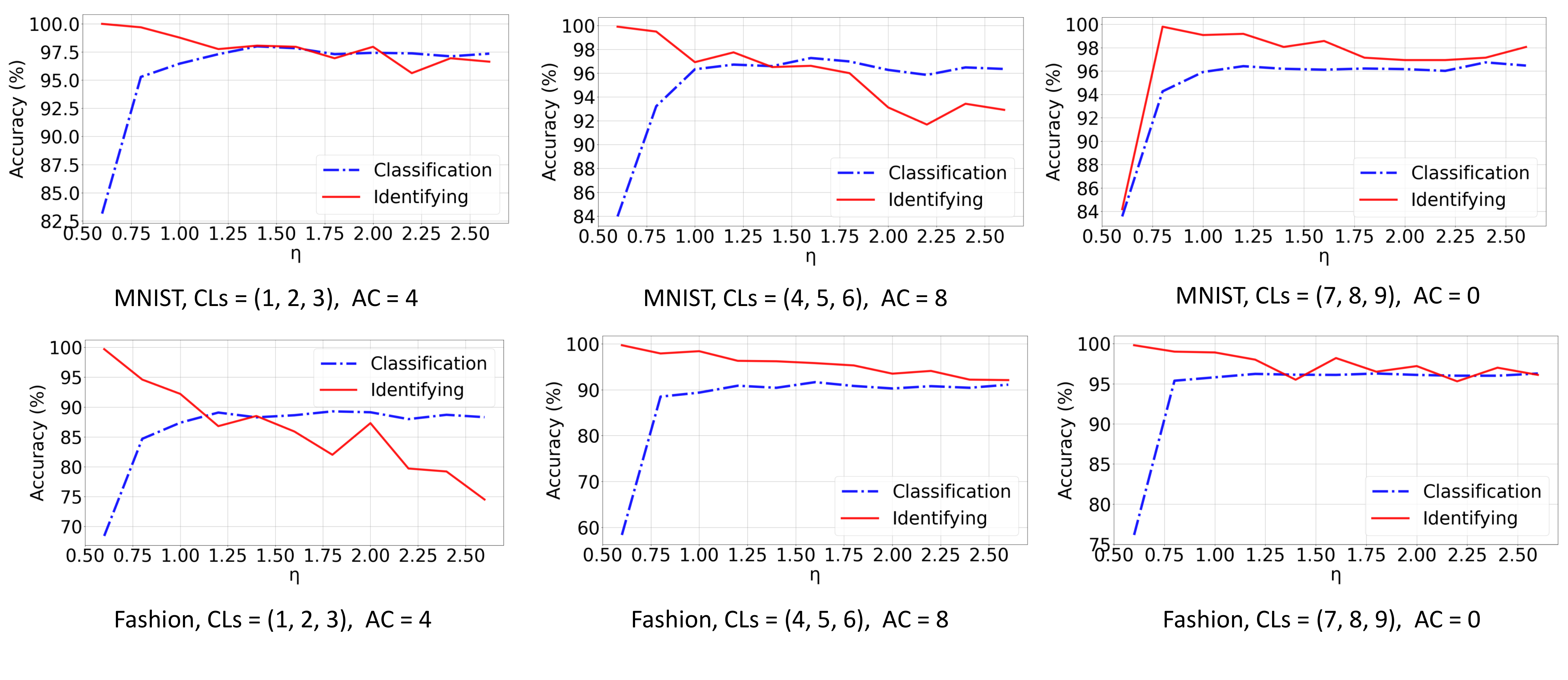}
	\caption{Illustrations classification accuracy for all classes and identifying accuracy for augmented classes  with various  perturbed mixture proportions. CLs $=(1, 2, 3)$ means that classes 1, 2, 3 are taken as complementary labels. AC $=4$ means that class 4 is taken as augmented class. $\eta$ denotes the  perturbed rate for mixture proportion.}
	\label{results_rate}
	\vspace{-0em}
\end{figure*}

\begin{figure*}
	\vspace{-0em}
	\centering
	\includegraphics[width=0.97\textwidth]{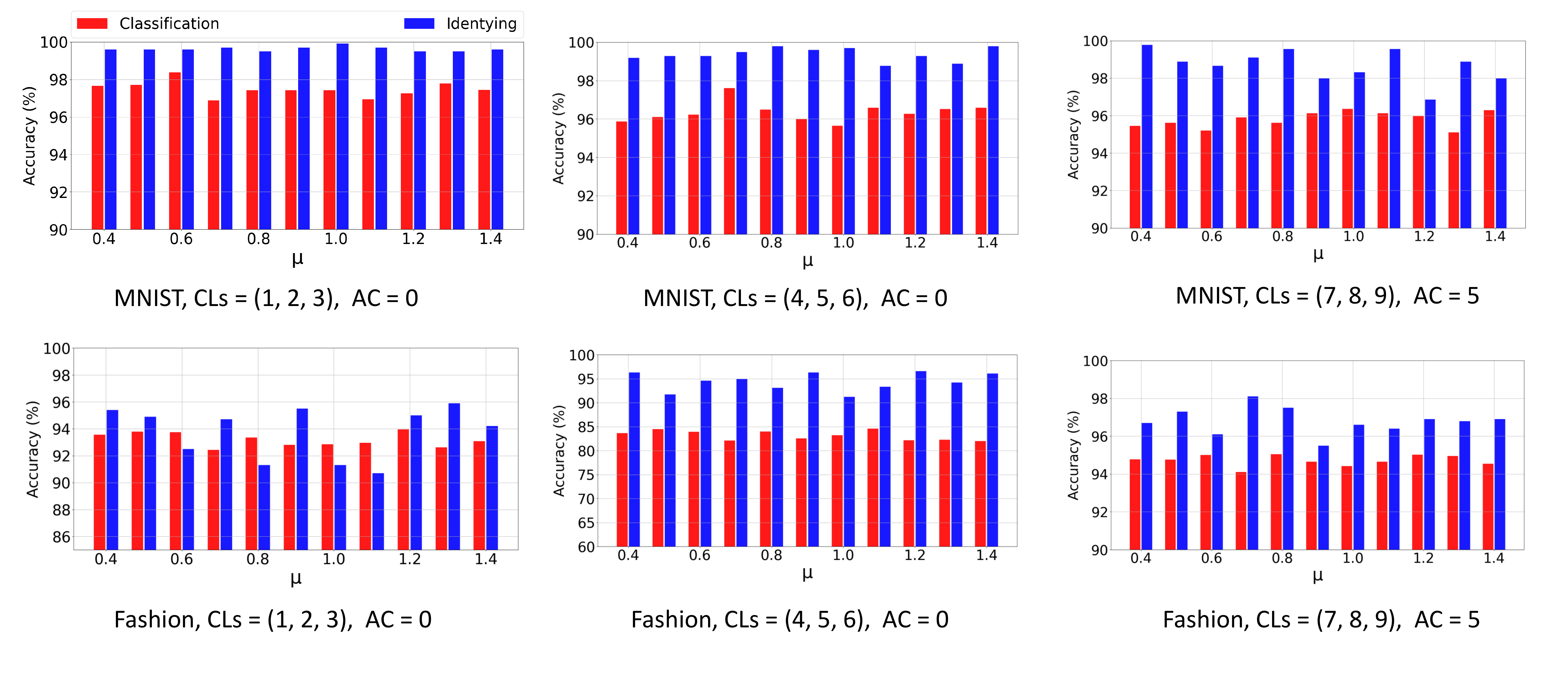}
	\caption{Illustrations classification accuracy for all classes and identifying accuracy for augmented classes  with various  perturbed rate of testing class distribution. CLs $=(1, 2, 3)$ means that classes 1, 2, 3 are taken as complementary labels. AC $=0$ means that class 0 is taken as augmented class. $\mu$ denotes the  perturbed rate of testing class distribution.}
	\label{results_dis}
	\vspace{-1em}
\end{figure*}

\textbf{Common Setup:} We conduct experiments using OVR strategy implemented by margin square loss $\phi (z) = (1-z)^2$. For  MNIST, Fasion-MNIST and  Kuzushi-MNIST dataset, we also use  neural network with 2 convolutional layers and 2 fully-connected layers as classifier. For CIFAR-10 dataset, we also use  neural network with 4 convolutional layers and 2 fully-connected layers as classifier. For \textbf{EUAC}  and \textbf{UREA}, the complementary labels dataset is treated as ordinary labels dataset and unlabeled dataset is utilized as augmented classes dataset. For \textbf{CL-PC} and \textbf{CL-GA}, they only use complementary labels dataset to train classifier and the unlabeled dataset is discarded in the training.
In this paper, we conduct our experiments on PyTorch and implementation on NVIDIA 3080Ti GPU.  We use Adadelta as optimizer with initial learning rate $5e^{-1}$ and batch size in 256 in our experiments.
\begin{table}
	\centering
	\vspace{-1em}
	\caption{Identifying accuracy of augmented classes on benchmark datasets, with varying classes and the number of examples. (CLs), A means that classes $(1, 2,3), 4 $ are taken as complementary labels and augmented classes respectively.  We report the mean and standard deviation of results over 3 trials. The best method is shown in bold (under 5$\%$ t-test). }
	{\resizebox{0.46\textwidth}{24mm}{
			\renewcommand\arraystretch{1.1}
			\tabcolsep=0.2cm
			\begin{tabular}{lc|ccc }
				\toprule
				Dataset&(CLs), A&EUAC&UREA&UCLLAC\\
				\cmidrule{1-5} 
				&(0,1,2), 3&77.62$\pm$11.22&67.32$\pm$13.33&\textbf{97.72$\pm$0.26}\\ 
				&(1,2,3), 5&53.73$\pm$3.65&38.52$\pm$4.69&\textbf{98.35$\pm$0.72}\\
				&(1,2,3), 9&49.35$\pm$1.24&45.85$\pm$12.39&\textbf{97.72$\pm$0.86}\\ 
				MNIST&(4,5,6), 2&53.42$\pm$13.22&57.52$\pm$2.42&\textbf{98.22$\pm$0.40}\\
				&(4,5,6), 3&63.76$\pm$12.43&62.27$\pm$9.86&\textbf{98.74$\pm$0.20}\\
				\cmidrule{1-5} 				
				&(1,2,3), 0&66.76$\pm$1.70&68.13$\pm$7.37&\textbf{92.56$\pm$1.51}\\ 
				&(1,2,3), 9&63.13$\pm$5.87&64.30$\pm$38.27&\textbf{99.80$\pm$0.10}\\
				&(1,2,3), 8&59.20$\pm$9.57&68.16$\pm$9.97&\textbf{98.80$\pm$0.21}\\ 
				Fashion&(4,5,6), 9&72.30$\pm$8.38&69.33$\pm$3.43&\textbf{99.10$\pm$0.86}\\
				&(4,5,6), 3&67.76$\pm$3.63&77.66$\pm$5.35&\textbf{93.60$\pm$3.21}\\
				\bottomrule
	\end{tabular}}}
	\vspace{-1.5em}
	\label{tab_iac_mnist}	
\end{table}

\subsection{Performance Comparison}
In this section, we  record  experimental results which demonstrate  performances of recognizing  known classes and classifying augmented classes in the testing dataset. Table.\ref{tab_acc_mnist} and Fig.\ref{results_cifar} report the recording results for MNIST, Fashion-MNIST,  Kuzushi-MNIST and CIFAR-10 datasets, which show that UCLLAC outperforms the other methods in most experiments. Since recent complementary labels methods can not classify the augmented classes, the results are not reported in this paper. As the results shown in Table.\ref{tab_iac_mnist}, the proposed method outperforms others which were proposed for identifying the augmented classes.  Besides, compared to the results of recognizing known classes, identifying augmented classes achieve better performances, which implies learning from complementary labels is more difficult than learning from augmented classes.

\subsection{Robustness for Inaccurate Training Class Priors \label{ritc}}
In the above setting, the  class prior is available in the training stage. In practice, given the complementary labels and unlabeled data, the mixture proportion can be estimated by Mixture Proportion Estimation (MPE) \cite{DBLP:conf/icml/RamaswamyST16}. Without loss of generality, we conduct the experiments for evaluating the robustness of inaccurate training class priors by varying degrees of inaccuracies in the training phase. Let $\eta$ be real number, $\vartheta  = \eta \theta$ be perturbed mixture proportions. In this section, the data is sampled by $ \theta$ but the classifiers are trained by using $\vartheta$ instead. Fig.\ref{results_rate} shows the experimental results, where the mixture proportion varies from 0.6 to 2.6 under the ground-true priors. We observe that the classification accuracy of proposed method is robust to inaccurate mixture proportion in mild environment. It is worth noting that the proposed method work well with $1< \eta <2.6$, which means that we can use a value $\vartheta_K$ ($\theta < \vartheta_K< 2\theta$) to train the classifiers, when the mixture proportion is missing. 

\subsection{Handling Class Probabilities Shift}
In above section, we have assumed that the distribution of  unlabeled data is same as the testing distribution, which is a ideal condition for learning from the augmented classes data. In this section, we conduct the experiments for investigating class distribution shift in the testing data. Without loss of generality, we evaluate our method on varying degrees of distribution shift for testing data. Let $\mu$ be real number, $p'_{te}(y) = \mu p_{te}(y)$ be perturbed distribution for testing data.  In the experiments,  the classifiers are trained by using $p_{te}(y)$ while tested by using $p'_{te}(y)$. We report experimental results in Fig.\ref{results_dis} by using neural networks on MNIST. From this Figure, we can find that equipped with unbiased risk, complementary labels learning with augmented classes works well for class probability shift. This observation is clearly in accordance with consistency analysis, as the method will obtain the Bayes classifiers, i.e, $f^*(x) = arg max_{c\in \{1,...K,K+1\}}P_{te}(y=c|x)$, where $P_{te}(y=c|x)$ denotes the conditional distribution for testing data.
\section{Conclusion}
In this paper, we investigate the complementary labels learning with augmented classes by exploring and exploiting unlabeled data.  We propose a novel problem setting called  Complementary Labels Learning with Augmented Classes (CLLAC), which brings the challenge that classifiers trained by complementary labels  should not only be able to  classify  the  instances from observed classes accurately, but also recognize the instance from augmented classes in the testing phase. Besides, we propose an unbiased risk estimator for complementary labels learning with augmented classes which  will obtain the Bayes classifiers over the testing distribution. Extensive experiments demonstrated that the proposed method outperforms  current state-of-the-art methods on several benchmark. 

In the future, we will study multi-complementary labels learning with unobserved augmented classes, which is a common setting in real-world environments. Besides, it  would be interesting to investigate the advanced method for complementary labels learning with augmented classes.

\appendix

This is the supplemental material for the paper  `` Complementary Labels Learning with Augmented Classes"

\section{B. Proof of Lemma 1}
Before showing the proof of Lemma 1 for self-contentedness, we introduce the rewritten classification risk of complementary labels as the following theorem.

\textbf{Theorem 6.} (Theorem 1 of \cite{DBLP:conf/icml/IshidaNMS19}) For any ordinary distribution $D$ and complementary distribution $\bar D$ related by 	$\overline p (x,\overline y ) = \frac{1}{{k - 1}}\sum\nolimits_{y \ne \overline y } {p(x,y)}$ with decision function $F$, and loss $l(y,F(x)) = \phi ({f_y}(x)) + \sum\nolimits_{k = 1,k \ne y}^{K} \phi ( - {f_k}(x))$, we have
\begin{equation}
	{R_\phi }(F) = \mathbb{E}_{(x,\bar y) \sim {{\bar p}}} [- (K - 1)l(\bar y,F(x)) + \sum_{j = 1}^K {l(j,F(x))}] 
\end{equation}

\emph{Proof of Lemma 1.}  First, recall that the risk with respect to OVR stragy is introduced as follows:
\begin{equation}
	\begin{split}
		{R_\phi }({f_1},...,{f_{K+1}}) &= \mathbb{E}_{(x,y) \sim {p_{te}}}[\phi ({f_y}(x)) \\ & + \sum\nolimits_{k = 1,k \ne y}^{K + 1} \phi ( - {f_k}(x))] 
	\end{split}
\end{equation}
Then, according to Theorem 6, the risk of complementary labels and augmented classes is formulated as follows: 
\begin{equation}\label{or_risk}
	\begin{split}
		{R_\phi } &= \mathbb{E}_{(x,y) \sim {p_{te}}}[\phi ({f_y}(x))  + \sum\nolimits_{k = 1,k \ne y}^{K + 1} \phi ( - {f_k}(x))] \\& = \theta \mathbb{E}_{(x,\bar y) \sim {{\bar p}_{kcl}}} [- (K - 1)l(\bar y,F(x)) + \sum_{j = 1}^K {l(j,F(x))}] \\& + (1 - \theta ){E_{(x,y) \sim {p_{ac}}}}[\phi ({f_y}) + \sum\limits_{i = 1,i \ne y}^{K + 1} {\phi ( - {f_i}(x))} ]
	\end{split}
\end{equation}

Since $p_{ac}(x,y\ne ac) =0$ hold for all $x \in \mathcal{X}$, we can reform the second term of Eq.\ref{or_risk} into
\begin{equation}\label{risk_ac}
	\begin{split}
		&\;\;\;\;(1 - \theta ){\mathbb{E}_{(x,y) \sim {p_{ac}}}}[\phi ({f_y}) + \sum\limits_{i = 1,i \ne y}^{K + 1} {\phi ( - {f_i}(x)} ]\\& = (1 - \theta ){\mathbb{E}_{x \sim {p_{ac}(x)}}}[\phi(f_{K+1}) + \sum\limits_{i = 1}^{K} {\phi ( - {f_i}(x)}) ]
	\end{split}
\end{equation}

Then, by summing over the label space $\mathcal{Y}$, the marginal density  is obtained as 
\begin{equation}
	(1 - \theta ){p_{ac}}(x) = {p_{te}}(x) - \theta {\bar p_{kcl}}(x)
\end{equation} where Eq.\ref{risk_ac} can be rewritten as follows:
\begin{equation}\label{lemma1}
	\begin{split}
		&\;\;\;\;(1 - \theta ){\mathbb{E}_{(x,y) \sim {p_{ac}}}}[\phi ({f_y}) + \sum\limits_{i = 1,i \ne y}^{K + 1} {\phi ( - {f_i}(x)} ]\\&=
		{{\mathbb{E}}_{{\rm{x}} \sim {p_{te}}(x)}}[l(K + 1,F(x))] - \theta {\mathbb{E}_{{\rm{x}} \sim {p_{kcl}}(x)}}[l(K + 1,F(x))]
	\end{split}
\end{equation} where $l(K + 1,F(x)) = \phi ({f_{K+1}}(x)) + \sum\nolimits_{i = 1}^{K} \phi ( - {f_i}(x))$

Plugging Eq.\ref{lemma1} into Eq.\ref{or_risk}, we complete the proof. \hfill $\square$

\section{B. Proof of Proposition 2}

\emph{Proof of Proposition 2.} According to Lemma 1, the risk can be rewritten as follows: 
\begin{equation} \label{prop2}
	\begin{split}
		R_{CLLAC}(F) & =  {{\theta \mathbb{E}}_{(x,\bar y) \sim {{\bar p}_{kcl}}}} \big [- (K - 1)l(\bar y,F(x)) \\& \;\;\;\; + \sum_{j = 1}^K {l(j,F(x))} \big ] + {{\mathbb{E}}_{{\rm{x}} \sim {p_{te}}(x)}}[l(K + 1,F(x))] \\& \;\;\;\;- \theta {\mathbb{E}_{{\rm{x}} \sim {p_{kcl}}(x)}}[l(K + 1,F(x))]\\& =  {{\theta \mathbb{E}}_{(x,\bar y) \sim {{\bar p}_{kcl}}}} \bigg [- (K - 1)l(\bar y,F(x)) \\& \;\;\;\;+ \sum_{j = 1}^K {l(j,F(x))} - l(K + 1,F(x)) \bigg ] \\& \;\;\;\;+ {{\mathbb{E}}_{{\rm{x}} \sim {p_{te}}(x)}}[l(K + 1,F(x))] 
	\end{split}
\end{equation}

Plugging 
\begin{equation}\label{loss}
	l(j,F(x)) = \phi ({f_j}) + \sum\limits_{k = 1,k \ne j}^{K + 1} {\phi ( - {f_k})} 
\end{equation} into Eq.\ref{prop2}, we complete the proof. \hfill $\square$

\section{B. Proof of Theorem 3}
In this section, we introduce the Infinite-Sample Consistency (ISC) of One-Versus-Rest (OVR) strategy provided by Zhang \cite{DBLP:journals/jmlr/Zhang04a} as follows:

\textbf{Theorem 7} (Theorem 10 of Zhang \cite{DBLP:journals/jmlr/Zhang04a}). Consider OVR method with margin surrogate loss, i.e., $l(y,F(x)) = \phi ({f_y}(x)) + \sum\nolimits_{k = 1,k \ne y}^{K+1} \phi ( - {f_k}(x))$. Assume $\phi$ is convex, differentiable,  bounded below and $\phi(z)<\phi(-z)$ when $z>0$. Then, OVR method is ISC on $\Omega = \mathbb{R}^{K+1}$ with respect to 0-1 classification risk.

The following results theorem in \cite{DBLP:journals/jmlr/Zhang04a} shows the relationship between the Bayes error and the ISC method risk.

\textbf{Theorem 8} (Theorem 3 of Zhang \cite{DBLP:journals/jmlr/Zhang04a}). Let $\mathcal{B}$ denotes the set of all vector Borel measurable functions, which take values in $ \mathbb{R}^{K+1}$. For $\Omega = \mathbb{R}^{K+1}$, let $\mathcal{B}_{\Omega} = \{F \in \mathcal{B}:\forall x, F(x) \in \Omega  \}$. If $[l(y, \bullet)]$ is ISC on $\Omega$ with respect to 0-1 classification risk, then for any $\epsilon_1 > 0$, there exists $\epsilon_2 > 0$ such that for all underlying Borel probability measurable $D$, and $F(\bullet) \in \mathcal{B}_{\Omega}$,
\begin{equation}
	\mathbb{E}_{(x,y) \sim D}[l(y,F(x))]  \leqslant \inf_{F' \in  \mathcal{B}_{\Omega}} \mathbb{E}_{(x,y) \sim D}[l(y, F'(x))] + \epsilon_2 
\end{equation} implies 
\begin{equation}
	R(T(F(x))) \leqslant R^* + \epsilon_1
\end{equation} where $T(F(x)) := argmax_{i=1,...,K+1}f_i(x)$, and $R^*$is the optimal Bayes error.

\emph{Proof of Theorem 3.} According to the lemma 1 and Proposition 2, the CLLAC risk can be rewritten by using complementary labels and augmented classes data distribution. Therefore, it is sufficient to demonstrate that consistency property is proven by using Theorem 7 and Theorem 8  for proposed method of CLLAC with OVR strategy. \hfill $\square$

\section{C. Proof of Lemma 4}
\textbf{Lemma 4.} For any $\delta>0$, with the probability at least $1-\delta/2$, we have
\begin{equation}
	\begin{split}
		{\sup _{F \in \mathcal{H}}}| & {{R_{kcl}}(F)-  {\widehat{R}_{kcl}}  (F)} | \\& \leqslant  4{K}{L_\phi }{\mathfrak{R}_{{n_{kcl}}}}(\mathcal{H}) + 2{K}{C_\phi }\sqrt {\frac{{\ln (4/\delta )}}{{2{n_{kcl}}}}} 
	\end{split}	
\end{equation}
\begin{equation}
	\begin{split}
		{\sup _{F \in \mathcal{H}}}&| {{R_{u}}(F) -  {\widehat{R}_{u}}  (F)} | \\& \leqslant 2{(K+1)}{L_\phi }{\mathfrak{R}_{{n_{kcl}}}}(\mathcal{H}) + {(K+1)}{C_\phi }\sqrt {\frac{{\ln (4/\delta )}}{{2{n_{u}}}}} 
	\end{split}
\end{equation} where $K$ denotes the number of known classes, $R_{kcl}(F) ={\mathbb{E}_{(x,\bar y) \sim {{\bar p}_{kcl}}}}[(\phi ( - {f_{K + 1}}) - \phi ({f_{K + 1}})) + (K - 1)[\phi ( - {f_{\bar y}}) - \phi ({f_{\bar y}})]]$ and $R_u(F) = {\mathbb{E}_{x \sim {p_{te}}(x)}}[\phi ({f_{K + 1}}) + \sum\limits_{j = 1}^K {\phi ( - {f_j})}] $, ${\widehat{R}_{kcl}}  (F)$ and  ${\widehat{R}_u}  (F)$ denote the empirical risk estimator to  $R_{kcl}(F)$ and $R_u(F)$ respectively, $\mathfrak{R}_{{n_{kcl}}}(\mathcal{H})$ and $\mathfrak{R}_{{n_u}}(\mathcal{H})$ are the Rademacher complexities \cite{DBLP:books/daglib/0034861} of $\mathcal{H}$ for the sampling of size $n_{kcl}$ from $\overline p_{kcl}(x,\overline y)$ and the sampling of size $n_u$ from $p_{te}(x,y)$. 

\emph{Proof of Lemma 4.} Suppose the surrogate loss $\phi(z)$ is bounded by $su{p_{z}}\phi(z)\leqslant C_{\phi}$, let function $\Phi_p$ defined for any complementary labels  sample $S_{kcl}$ by $\Phi (S_{kcl}) = sup_{F\in \mathcal{H}} R_{kcl}( {F}) -  \widehat R_{kcl}( {F} )$. If  $x_i$ in complementary labels dataset is replaced with $x_i'$, the change of $\Phi_{kcl} (S_{kcl})$ does not exceed the supermum of the difference, we have
\begin{equation}
	\Phi_{kcl} (S_{kcl}') - \Phi_{kcl} (S_{kcl}) \leqslant 2\frac{{{C_\phi }{K}}}{{{n_{kcl}}}}
\end{equation} Then, by McDiarmid's inequality, for any $\delta > 0 $, with probability at least $1-\delta/2$, the following holds:
\begin{equation}
	\begin{split}
		sup_{F\in \mathcal{H}} &| \widehat R_{kcl}( {F} ) - R_{kcl}( {F})| \\& \leqslant \mathbb{E}_{S_{kcl}} \Phi_{kcl} (S_{kcl}) + 2{K}{C_\phi }\sqrt {\frac{{\ln (4/\delta )}}{{2{n_{kcl}}}}}
	\end{split}
\end{equation}.

By using the Rademacher complexity \cite{DBLP:books/daglib/0034861}, we can obtain 
\begin{equation}
	\begin{split}
		sup_{F\in \mathcal{F}} | \widehat R_{kcl}( {F} ) - R_{kcl}( {F})| \leqslant 2 \mathfrak{R}_{{n_{kcl}}}({ \widetilde l_{kcl} }{\circ \mathcal H}) \\ + 2{K}{C_\phi }\sqrt {\frac{{\ln (4/\delta )}}{{2{n_{kcl}}}}}
	\end{split}
\end{equation} where $\mathfrak{R}_{{n_{kcl}}}({ \widetilde l_{kcl} }{\circ \mathcal H})$ is the Rademacher complexity of the composite function class (${ \widetilde l_{kcl} }{\circ \mathcal H}$) for examples size $n_{kcl}$. As $L_\phi$ is the Lipschitz constant of $\phi$, we have $\mathfrak{R}_{{n_{kcl}}}({ \widetilde l_{kcl} }{\circ \mathcal H}) \leqslant 2KL_\phi {\mathfrak{R}_{{n_{kcl}}}}(\mathcal{H})$ by Talagrand's contraction Lemma \cite{DBLP:books/daglib/0034861}, where $\widetilde l_{kcl}(\bar y,F(X)) = \phi ( - {f_{K + 1}}) - \phi ({f_{K + 1}})) + (K - 1)[\phi ( - {f_{\bar y}}) - \phi ({f_{\bar y}})]$. Then, we can obtain the 
\begin{equation}
	\begin{split}
		{\sup _{F\in \mathcal{H}}}\left| {{R_{kcl}}(F) -  {\widehat{R}_{kcl}}  (F)} \right| \leqslant 4{K}{L_\phi }{\mathfrak{R}_{{n_{kcl}}}}(\mathcal{H}) \\ + 2{K}{C_\phi }\sqrt {\frac{{\ln (4/\delta )}}{{2{n_{kcl}}}}} 
	\end{split}
\end{equation}

${\sup _{F \in \mathcal{H}}}\left| {{R_u}(F) -  {\widehat{R}_u}  (F)} \right|$ can be proven using the same proof technique, which proves Lemma 4. \hfill $\square$

\section{D. Proof of Theorem 5}
\textbf{Theorem 5.} For any $\delta>0$, with the probability at least $1-\delta/2$, we have
\begin{equation*}
	\begin{split}
		R_{CLLAC}&({\hat {F}_{CLLAC})} -  \mathop{\rm {min}} _{{F} \in \mathcal{H}}R_{CLLAC}(F) \leqslant  \\& 8{K}{L_\phi }{\mathfrak{R}_{{n_{kcl}}}}(\mathcal{H}) + 4{(K+1)}{L_\phi }{\mathfrak{R}_{{n_u}}}(\mathcal{H})\\&+ 4{K}{C_\phi }\sqrt {\frac{{\ln (4/\delta )}}{{2{n_{kcl}}}}}+ 2{(K+1)}{C_\phi }\sqrt {\frac{{\ln (4/\delta )}}{{2{n_u}}}}
	\end{split}
\end{equation*} where $\hat {F}_{CLLAC}$ is trained by minimizing the CLLAC risk $R_{CLLAC}$

\emph{Proof of Theorem 5.}
According to Lemma 4, the estimation error bound is proven through 
\begin{align*}
	R_{CLLAC}&({\widehat {F}_{CLLAC}}) - R_{CLLAC}({\mathbf F^*})   \\&= ( \widehat R_{CLLAC}({\widehat {F}_{CLLAC} }) - \widehat R_{CLLAC}({\widehat {F^*}})) \\& \; \; \; + (R_{CLLAC}({\widehat {F}_{CLLAC}}) - \widehat R_{CLLAC}({\widehat { F}_{CLLAC}})) \\& \; \; \; + (\widehat R_{CLLAC}({\widehat{F^*} }) - R_{CLLAC}({\widehat {F^*}}))
	\\& \leqslant 0 + 2 sup_{F\in \mathcal{H}} | R_{CLLAC}( {F}) - \widehat R_{CLLAC}( {F} ) |
\end{align*} where $\mathbf F^* = arg\mathop{\rm {min}}_{{F} \in \mathcal{H}}R(F)$.

We have seen the definition of $ R_{CLLAC}({ {F}}) $  and $ \widehat R_{CLLAC}({ {F}}) $ that can also be decomposed into $		{R_{CLLAC}(F)} = \theta {\mathbb{E}_{(x,\bar y) \sim {{\bar p}_{kcl}}}}[(\phi ( - {f_{K + 1}}) - \phi ({f_{K + 1}})) + (K - 1)[\phi ( - {f_{\bar y}}) - \phi ({f_{\bar y}})]] + {\mathbb{E}_{x \sim {p_{te}}(x)}}[\phi ({f_{K + 1}}) + \sum\limits_{j = 1}^K {\phi ( - {f_j})} ] $and $ 		{\hat R_{CLLAC}} =  \frac{\theta}{{{n_{kcl}}}} \sum\limits_{i = 1}^{{n_{kcl}}} [(\phi ( - {f_{K + 1}}({x_i})) - \phi ({f_{K + 1}}({x_i}))) + (K - 1)[\phi ( - {f_{{{\bar y}_i}}}({x_i})) - \phi ({f_{{{\bar y}_i}}}({x_i}))]]  +  \frac{1}{{{n_{u}}}}\sum\limits_{i = 1}^{{n_u}} {[\phi ({f_{K + 1}}({x_i})) + \sum\limits_{j = 1}^K {\phi ( - {f_j}({x_i}))} ]}  $.  Due to the sub-additivity of  the supremum operators with respect to risk, it holds that 
\begin{equation*}
	\begin{split}
		sup_{F\in \mathcal{H}} |\widehat R_{CLLAC}( {F} ) - R( {F}) |  \leqslant  sup_{F\in \mathcal{H}} | \widehat R_{kcl}( {F} ) - R_{kcl}( {F})|  \\ +  sup_{F\in \mathcal{H}} |\widehat R_{u}( {F} ) - R_u( {F})| 
	\end{split}
\end{equation*}

According to the Lemma 4, we can get the generalization bound that 
\begin{equation*}
	\begin{split}
		R_{CLLAC}&({\hat {F}_{CLLAC})} - \mathop{\rm {min}} _{{F} \in \mathcal{H}}R_{l_{CLLAC}}(F) \\& 8{K}{L_\phi }{\mathfrak{R}_{{n_{kcl}}}}(\mathcal{F}) + 4{(K+1)}{L_\phi }{\mathfrak{R}_{{n_u}}}(\mathcal{H}) \\& + 4{K}{C_\phi }\sqrt {\frac{{\ln (4/\delta )}}{{2{n_{kcl}}}}} + 2{(K+1)}{C_\phi }\sqrt {\frac{{\ln (4/\delta )}}{{2{n_u}}}}
	\end{split}
\end{equation*} with probability at least $1-\delta/2$, which finishes the proof. \hfill $\square$





\bibliographystyle{elsarticle-num} 
\bibliography{elsarticle-template-num}
\end{document}